\documentclass{article}
\usepackage[T1]{fontenc}
\usepackage[utf8]{inputenc}
\usepackage{lmodern}
\usepackage[hscale=0.7,vscale=0.8]{geometry}
\usepackage[english]{babel}
\usepackage{csquotes}
\usepackage{authblk}
\usepackage{booktabs} 
\usepackage[backend=biber]{biblatex}
\bibliography{main}
 \usepackage{graphicx} 
 \usepackage{pifont}
\usepackage{multirow}
\begin{document}
\title{\textbf{A Perspective on Individualized Treatment Effects Estimation
from Time-series Health Data}}
\author[1]{Ghadeer O. Ghosheh}
\author[1]{Moritz Gögl}
\author[1]{Tingting Zhu}
\affil[1]{Institute of Biomedical Engineering, University of Oxford}
\date{}

\maketitle
\section*{Abstract}
The burden of diseases is rising worldwide, with unequal treatment efficacy for patient populations that are underrepresented in clinical trials. Healthcare, however, is driven by the average population effect of medical treatments and, therefore, operates in a “one-size-fits-all” approach, not necessarily what best fits each patient. These facts suggest a pressing need for methodologies to study individualized treatment effects (ITE) to drive personalized treatment. Despite the increased interest in machine-learning-driven ITE estimation models, the vast majority focus on tabular data with limited review and understanding of methodologies proposed for time-series electronic health records (EHRs). To this end, this work provides an overview of ITE works for time-series data and insights into future research. The work summarizes the latest work in the literature and reviews it in light of theoretical assumptions, types of treatment settings, and computational frameworks. Furthermore, this work discusses challenges and future research directions for ITEs in a time-series setting. We hope this work opens new directions and serves as a resource for understanding one of the exciting yet under-studied research areas.

\section{Introduction}
\label{ITEs}
Medical treatment and drug budgets are the highest burdens on governments and medical institutions. Despite these high costs, only about 90\% of drugs work for 30-50\% of the population \cite{the2018personalised}. These statistics suggest a pressing need to identify patient subgroups where personalized treatments can be prescribed. Studying treatment effects has gained much attention over the past years, where various tools and approaches have been proposed to help mitigate the financial cost and optimize the effectiveness and efficacy of prescribed treatments. One of the fast-growing applications in clinical Machine Learning (ML) is studying Individualized Treatment Effects (ITEs)~\cite{bica2021real}, where ML capabilities, in many cases, have superseded those of state-of-the-art clinical and pharmaceutical advancements. Using the wealth of observational Electronic Health Records (EHRs) data, individualized patient response to various treatments can be estimated~\cite{bica2021real}.

While there has been increased attention to ITE works from static observational EHR data~\cite{yoon2018ganite,chipman2010bart}, much less attention has been given to those from EHR-data measured over time, often referred to as time-series EHRs. Despite similarities in the concepts for treatment estimation between static and time-series data, challenges related to the time-varying nature of covariates make many existing works not directly applicable to time-series data. To this end, we aim to provide an overview of the main concepts and ideas for estimating ITE from time-series data. Starting with introducing the concepts of Randomized Controlled Trials (RCTs) and observational data, then discussing topics such as efficacy and effectiveness of treatments and challenges in ITE estimation. This work aims to bridge the theoretical assumptions from causal inference and machine learning works to real-world EHR implications. To the best of our knowledge, this is the first piece of work comprehensively reviewing challenges and methods in ITE estimation for time-series data. We summarize the latest work in the literature and group them based on theoretical assumptions, types of treatment settings and computational frameworks. Lastly, this work discusses challenges and future research directions for ITEs in a time-series setting. We hope this work opens new directions and serves as a resource for understanding one of the exciting yet under-studied research areas.

\subsection{Randomized Controlled Trial Data}
RCT data are considered the gold standard for studying treatment effects. This mainly stems from the random assignment of participants to treatment groups, which eliminates confounding bias effectively. The randomness and control measures employed in RCTs make them a lucrative option for studying the effect of treatments; an unbiased estimate of the average treatment effects (ATE) can be directly computed from the data~\cite{lunceford2004stratification}.      
While RCTs offer methodological strengths, various challenges hinder their full and optimal usage for studying treatment effects~\cite{black1996we}. Firstly, despite the power of "randomness" to eliminate confounding, certain biases may remain in RCTs. This bias does not come from the treatment assignment level but from representatives of samples presented in the study, denoted as sample selection bias hereafter. Most RCTs tend to employ stringent exclusion criteria of enrolled participants, which might introduce bias in the results for the unrepresented population members~\cite{lindsay2020age,kim2010sex, schulz1995empirical, black1996we, stiller1994centralised}. This becomes a bigger issue when a significant percentage of treated patients in real-world data belong to the population excluded from the RCT. For example, the ageing population is rarely enrolled in diabetes RCTs due to their old age and multi-morbid health conditions, despite constituting a large proportion of diabetic patients~\cite{kalyani2017diabetes}. Such factors might introduce bias that makes RCTs' generalizability and external validity questionable~\cite{hayes1999simple,black1996we}.

The generalizability of RCT results is further limited by the fact that current RCTs are typically designed to only measure the ATE. In other words, they estimate how the "average patient" will respond to a given treatment. For a more personalized approach, the patient's unique characteristics would be needed to predict an individual patient's response to a given treatment, and it may differ significantly from the average response of a population. Furthermore, RCTs tend to be financially costly to design and implement \cite{horn2005another,sanson2007limitations} and their data tend to be hard to share due to privacy concerns~\cite{schelvis2015evaluation}. Additional limitations exist in terms of their relatively small sample size~\cite{hayes1999simple},  ethical issues~\cite{goldstein2018ethical}, and short lengths of follow-ups which might miss out on the long-term effects of medications~\cite{black1996we}. For example, the effects of oral contraceptives were not quantified until the presence of long-term data, which were not captured in RCTs~\cite{shaw1987adverse}.
        
\subsection{Observational Data: From Efficacy to Effectiveness}
Despite all of the stated challenges, RCTs are still essential for determining the \textit{Efficacy} for treatments~\cite{kraemer2000pitfalls} but are not necessarily optimal for studying treatment \textit{Effectiveness}. Making a clear distinction between these two terms will help one to understand when data-driven and statistical approaches can improve the use of RCT data. Efficacy refers to the effect of interventions under ideal "theoretical" circumstances, while Effectiveness means an effect is detected not under ideal but under real-world conditions~\cite{kraemer2000pitfalls}.
Observational EHR data presents a good candidate to better test for treatment effectiveness. Specifically, longitudinal observational data collected in EHRs typically includes a diverse patient cohort with no strict exclusion criteria, making it more representative of the real or targeted patient population. Observational data is also less expensive to collect when compared to RCTs and captures long-term outcomes~\cite{black1996we,newsome2018estimating}. Moreover, with the widespread use of EHR systems worldwide, observational data can allow for estimating treatment effects from diverse clinical settings such as low-middle-income countries (LMICs), where performing RCTs would not be feasible due to high costs. To this end, observational data is a promising resource for studying treatment effects with more inclusive estimations for various patient groups while maintaining low cost and learning from real-world evidence. 


\subsection{Challenges of Treatment Effects Estimation}

Treatment effect estimation is a subfield of causal inference, and as such suffers from the fundamental problem that counterfactual outcomes are never observed ~\cite{rosenbaum1983central}. A counterfactual outcome refers to the hypothetical outcome that would have been observed if a different treatment than the actual (factual) one had been given~\cite{bica2021real}. Of course, for treatments that were not administered, it is not possible to extract the ground truths of individual patients' outcomes directly, whether from observational or clinical trial data. In RCTs, this problem is resolved by estimating the ATE across the entire study population but not the ITE of an individual. Randomization of treatment allocation ensures that the underlying distribution of the patient characteristics is similar in both the treatment and control groups. This allows us to compare the average outcomes in both groups and thus determine the ATE.

Omitting the randomized control measures used in RCTs and relying on observational data precludes the direct computation of treatment effects. This is because the treatment assignment is not random, but biased, as treatment selection in observational data is often driven by the patient's characteristics, such as treatment allocation flowcharts in clinical practice guidelines~\cite{tobe2013clinical}. Therefore, clinical practice recorded in real-world observational data results in systematic differences in the characteristics of treated and untreated patients. To be able to estimate treatment effects from observational data, it is essential to remove the confounding bias, introduced by the non-random treatment assignment. A confounder is a variable that influences both the intervention and the outcome, potentially leading to a spurious association between them~\cite{anoke2019approaches}. For example, in clinical practice, patients with more severe health conditions might be more likely to be given stronger medications while still being expected to have poorer outcomes. However, it would be wrong to conclude that stronger medication leads to poorer patient outcomes. Failing to adjust for the severity of a health condition as a confounding factor in this case would lead to measuring a spurious association that does not reflect the actual treatment effect.
        

\subsection{From Average to Individualized Treatment Effects Estimation}
The modern understanding of estimating treatment effects is highly attributed to the work of Neyman-Rubin's "potential outcomes framework" ~\cite{rubin2005causal}. In the Neyman-Rubin model, the ITE between a treatment $A$ and a treatment $B$ is defined as the difference between the two potential outcomes (e.g., blood pressure) after administering a treatment $A$ or $B$ to a given patient. Subject to certain assumptions~\cite{bica2021real}, the ATE can be computed directly from RCT data by calculating the difference between the average outcomes in both treatment groups. However, due to the absence of counterfactuals in real-world data, ITEs cannot be calculated directly but must be estimated through the use of models. 
Based on statistical methods, but also driven by recent developments in machine and deep learning approaches, various models have been proposed to estimate treatment effects on a personalized level~\cite{yoon2018ganite,ghosh2021propensity,bica2020estimating,melnychuk2022causal}. Overall, most of these models estimate the potential outcome for an individual patient by learning the underlying effects and interactions between patients' characteristics, treatment and outcome. The patient's ITE can then be calculated as the difference between the predicted potential outcomes with and without treatment. Furthermore, various methods were proposed to address the problem of confounding bias introduced in observational studies as a result of unobserved confounders~\cite{robins2000marginal,bica2020time,kuzmanovic2021deconfounding}, paving
the way for ITE estimations for personalized medicine.

\section{Estimating Individualized Treatment Effects from Time-series Data}
Various works for estimating ITE using observational EHR data have been proposed recently~\cite{yoon2018ganite, shalit2017estimating,ghosh2021propensity}. Despite the plethora of works, most of them estimate the treatment effect using static data, where each patient is represented as a snapshot of covariates at the exposure of the treatment. While useful, using only static data and disregarding the time component have many limitations for estimating the impact of treatment over time. Furthermore, the ITE estimation from static data limits the opportunities for learning when to stop or change the treatments when the outcome is dynamic and varies over time, a critical clinical application for ITE. Additionally, in the static treatment effect estimation setup, the treatments are assigned at a single time point and often remain static over time. On the other hand, using time-series data for treatment effect estimation would allow for monitoring and changing the treatment plan dynamically while simultaneously observing the effect of time-varying treatment on patient covariates and outcomes of interest. A typical example of a time-series ITE problem is in the cancer application, where a patient's treatment option (e.g., radiation or chemotherapy) is adjusted according to his or her clinical response  (tumor size) over time. In the static setting, such a problem cannot be addressed due to the absence of the time factor.

\begin{figure}[h]
    \centering
    \includegraphics[width=\textwidth]{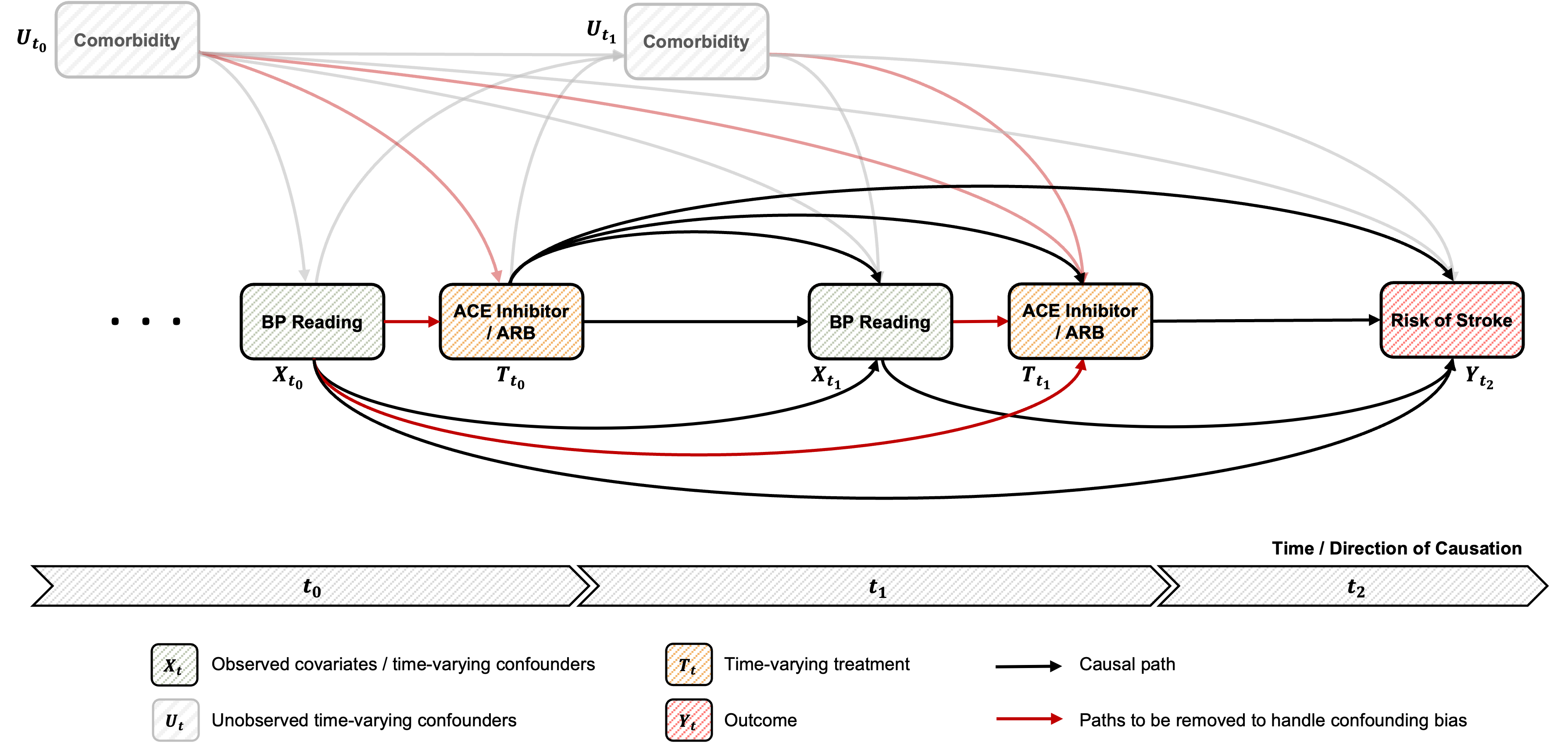}
    \caption{Illustration of a causal model underlying a dynamic ITE estimation setting with time-varying treatments and covariates. The arrows indicate causal dependencies between variables. Here, the observed covariates are blood pressure (BP) readings that act as time-varying confounders affecting all subsequent treatment decisions (ACE inhibitor vs. ARB) and the outcome of interest (risk of stroke). We also showcase an example of an unobserved "hidden" confounder, namely potential comorbidity that is not directly represented in the data. All connections from covariates and unobserved variables to the treatment variables are depicted in red, and would need to be accounted for to remove confounding bias effectively.}
    \label{fig:enter-label}
\end{figure}

Despite the promise and potential of treatment effect estimation from time-series data, the major problem lies in the time-varying or temporal confounders. Similar to the static confounders, a temporal confounder is typically a time-varying variable that impacts both the treatment assignment and outcome. For example, consider that Angiotensin receptor blockers (ARBs) (treatment 1) are given when a hypertension patient's blood pressure (covariate) is outside the normal range value. Suppose also that this patient's covariate was affected by the past administration of ACE inhibitor (treatment 0), another type of treatment for uncontrolled blood pressure. Estimating the effect of a different sequence of treatments on the patient outcome would require adjusting for the bias at the current step (treatment 1) and the bias introduced by the previous application of ACE inhibitors (treatment 0). Adjusting for time-varying confounders remains a major challenge hindering the direct application of methodologies developed for static treatment effects tasks to dynamic problem settings. Here we have reviewed works in the literature for estimating treatment effects from time-series data, including the estimation frameworks, model architectures and assumptions used. An overview of the existing work on ITE from time-series data is presented in Table~\ref{tab:summary}. They are categorized into two main groups: (i) outcome estimation methods which focus on inferring the ITE by estimating the potential outcomes of different treatments, and (ii) deconfounder methods, which estimate the ITE in the presence of hidden confounders. More details are presented in Sections~\ref{outcomeIte} and~\ref{deconfound}, covering ITE outcome estimation and deconfounder methods for time-series data, respectively. In Figure~\ref{fig:enter-label}, we show an example causal model that underlies a dynamic ITE estimation setting with time-varying treatments and covariates.

\begin{table}[ht!]
    \centering
        \caption{A summary of ITE works for time-series data.}
\resizebox{\textwidth}{!}{
    \begin{tabular}{llcccc}
    \toprule
        &
        \textbf{Proposed Methods} & \textbf{Estimation Framework} & \textbf{Assumptions} &\textbf{Model Architecture} &
        \textbf{Validation Data}\\ 
        \midrule
         \multirow{5}{*}{\textbf{Outcome Estimation Methods}}& MSM~\cite{robins2000marginal} & MSM & C/P/SSI  & LR & NA\\
          & RMSN ~\cite{lim2018forecasting} &  MSM& C/P/SSI   & LSTM & simulated tumor dynamics data \\
       & CRN~\cite{bica2020estimating}  &Balanced Representation& C/P/SSI  & LSTM & simulated tumor dynamics data, ICU data\\
         & G-Net~\cite{li2021g} &   G-formula& C/P/SSI& LSTM & \begin{tabular}[t]{@{}c@{}} simulated tumor dynamics data, \\ simulated cardiovascular data \end{tabular}\\
         & Causal Transformer ~\cite{melnychuk2022causal}  & Balanced Representation& C/P/SSI   & Transformer & \begin{tabular}[t]{@{}c@{}} simulated tumor dynamics data, \\ (semi-synthetic) ICU data \end{tabular}\\
         \midrule
        \multirow{3}{*}{\textbf{Deconfounder Methods}}&Time Series Deconfounder \cite{bica2020time} & Latent Factor Model& C/P/SSSI & RNN Factor Model &  simulated data, ICU data\\ 
         & Sequential Deconfounder \cite{hatt2021sequential} & Latent Factor Model& C/P/TIUC& GPLVM & simulated data, ICU data\\
         & Deconfounding Temporal AutoEncoder ~\cite{kuzmanovic2021deconfounding} & Noisy Proxies & C/P & AutoEncoder & simulated data, ICU data\\
         \bottomrule
    \end{tabular}}
    \label{tab:summary}
    \footnotesize{*The abbreviations in full form.  (C): Consistency, (SO): Sequential Overlap, (SSI): Sequential Strong Ignorability, (SSSI): Sequential Single Strong Ignorability, (TIUC): Time-Invariant Unobserved Confounding, (LR): Logistic Regression, (LSTM): Long Short-Term Memory, (RNN): Recurrent Neural Network, (GPLVM): Gaussian Process Latent Variable Model.}\\

\end{table}

\subsection{Assumptions}
Estimating treatment effects from time-series data relies on the potential outcome framework~\cite{rubin2005causal} and its extensions to the time-varying setting~\cite{robins2008estimation}. In the potential outcome framework, an ITE is the difference between the potential outcomes for a specific individual given different treatments.
Three main assumptions are typically required for treatment effects to be identifiable from time-series data. We explain each in lay terms and provide examples from real-world clinical problems. For interested readers, we provide references to works that explain and include mathematical notations. 
\begin{enumerate}

\item \textbf{Consistency.} 
This assumption states that the potential outcome of a patient should be consistent with his/her "factual" outcome if the same treatment plan is applied. For example, consider a patient who has been on a diabetes treatment plan $A$ for several years, where blood glucose is considered an outcome monitored over time. If a model is built to estimate this patient's potential outcome given the same treatment plan $A$, such outcome would be equivalent to the outcome observed for that patient (i.e., blood glucose).  The mathematical notation and theoretical basis for this consistency are explained 
in~\cite{bica2020estimating,melnychuk2022causal}.
\item{\textbf{Sequential Overlap or Positivity}.} This assumption means that each treatment option has a non-zero probability of being given to the patient at each timestep. For example, let us consider a cancer treatment where options are either radiotherapy or chemotherapy. If a patient was given radiotherapy last month, a doctor might give this patient chemotherapy or radiotherapy this month, and both have a non-zero probability of being given to the patient. 
\item \textbf{Sequential Strong Ignorability} This assumption means that conditioned on the observed patient history, the current treatment assignment is independent of the potential outcome. In some works, this assumption is referred to as sequential exchangeability or "no unobserved confounders".  In other words, no unobserved confounders affect both treatment and outcome. While useful, this assumption can't be tested in practice since various factors may impact the treatment and the outcome, yet they might not be recorded or observed. For this purpose, various works have been proposed to relax this assumption (see referenced works for estimation with hidden confounders described in Section ~\ref{deconfound}). Example variants include \textit{Sequential Single Strong Ignorability}, where the assumption is limited to no hidden single cause confounders~\cite{bica2020time}. Another example is \textit{Time-Invariant Unobserved Confounding}, where confounders exist with the condition that they are the same random variable at each time step~\cite{hatt2021sequential}.   
\end{enumerate}

\subsection{ITE Outcome Estimation Methods in time-series data}
\label{outcomeIte}
\subsubsection{Marginal Structural Models (MSM)}
Various approaches from epidemiology have been proposed for accounting for time-varying confounders, one of which is \textit{inverse probability of treatment weighting} (IPTW)~\cite{chesnaye2022introduction}. The main idea behind IPTW involves assigning weights that will redistribute or balance the population such that the effect of time-varying confounding is removed. The weights are derived based on the inverse probability of receiving the patient's treatment at each respective time point conditional on their covariate's history. The IPTW setup creates a pseudo population set where getting treatment assignment is independent of the underlying patient characteristics and previous treatment assignments~\cite{austin2015moving}. One way to implement IPTW in the time-series setting is via \textit{marginal structural models} (MSMs)~\cite{robins2000marginal}. MSMs focus on controlling the effects of time-varying confounders affected by previous treatment exposure. The name "Marginal" refers to the approach of estimating the marginal distribution of the treatment over time with respect to the outcome~\cite{williamson2017marginal}. Similarly, the name "structural" refers to the approach of causal relationship exploration inspired by econometrics~\cite{williamson2017marginal}. An MSM first calculates the weights, most commonly via a regression-based IPTW model, and assigns such weights to each observation. The estimated weights indicate whether each of the observations in confounders is under-represented or over-represented in the sample for a target population
~\cite{williamson2017marginal}. The use of sample reweighing aims to remove the imbalance and bias caused by the uneven distribution of time-varying confounders across treatment groups. Finally, the treatment effect can be estimated using the calculated weights. 

While powerful and useful, MSMs have limitations when dealing with high-dimensional and complicated data dynamics. This is because the treatment effects predictions are calculated using linear or logistic regression models. To address this, ~\cite{lim2018forecasting} proposed \textit{recurrent marginal structural networks} (RMSMs) where recurrent neural networks~\cite{connor1994recurrent} were used to estimate the inverse probability of treatment weights and the counterfactual treatment outcomes. Similar to the standard MSM two-stage approach, RMSM has two main networks. The first network calculates the treatment probability weights used for the IPTW. The second network, on the other hand, is the prediction used to determine the treatment response given a sequence of treatments and the calculated weights~\cite{lim2018forecasting}.
Despite the promise shown by the statistical and deep learning approaches, MSMs can be unstable if the IPTW results in extreme weights, leading to model misspecification~\cite{bica2020estimating}.
\subsubsection{G-formula}
Another method for estimating treatment effects in time-varying confounders is \textit{G-formula}. G-formula was first described by Robins et al.~\cite{robins1986new}, where the author proposed a method for generalization of standardization to time-varying treatments and confounders and referred to it as the G-computation algorithm formula. Most of the works from epidemiology use the term G-formula or G-computation to refer to the same method proposed by~\cite{robins1986new}. The key assumption for the G-computation formula is that the treatment received at each time was allocated conditional on the observed past treatment and covariate history~\cite{robins1986new}. G-formula works by estimating the conditional distribution of relevant covariates given covariate and treatment history at each time point, then producing Monte Carlo estimates of counterfactual outcomes by simulating forward patient trajectories under treatment strategies of interest~\cite{naimi2017introduction}. In most statistical works, the estimation of patient trajectories and outcomes is done via simplistic regression estimators. While useful, it is important to remember that simple regression models fail to capture complex dependencies over time when dealing with high-dimensional time-varying data. In terms of implementation, there are no well-established G-formula implementations in statistical packages, which limits its applicability when compared to MSMs. Recently, ~\cite{li2021g} proposed G-NEt, the first deep-learning work that estimates ITEs via an LSTM-based G-formula model. The G-Net results showed improved performance when compared to those estimated using a logistic regression estimator and other deep-learning-based models~\cite{lim2018forecasting,bica2020estimating}.
\subsubsection{Balanced Representations}
Unlike MSM and G-formula-based estimations, a new class of estimation from deep learning evolved based on learning representations that balance the distribution of the treatment and control groups. The original works for learning balanced representations were first proposed for static settings~\cite{johansson2016learning}, several studies then used deep learning architecture to learn treatment invariant representation for each time step to remove the association between the patient history and treatment assignment. For example, the \textit{counterfactual recurrent network} (CRN)~\cite{bica2020estimating} is the first work to use a sequence-to-sequence model to learn balanced representations via adversarial training. In their proposed work, CRN aims to learn representations not predictive of treatment assignments yet achieve the highest performance in predicting the outcome.  Another related work is that of ~\cite{melnychuk2022causal}, where the authors proposed \textit{Causal Transformer}, which is a transformer-based model that aims to learn treatment invariant balanced representations to estimate ITE over time.  To do so, the Causal Transformer comprises three transformer sub-networks for processing the time-varying covariates, treatments and outcomes, all of which are combined via a joint network with cross-attentions.

\subsection{ITE Deconfounding methods for time-series data}
\label{deconfound}
\subsubsection{Latent Factor Models}
All of the aforementioned studies for estimating ITE in time-series data focus on settings where all confounders are observed, or in other words; they require sequential strong ignorability assumption to hold. Despite the potential of such works, the sequential ignorability is not testable in practice. To this end, several studies have proposed approaches where sequential ignorability is relaxed to account for settings where forms of hidden confounders exist in the data. For example, the first work to propose a deep learning model for deconfounding time-series data was the \textit{Time Series Deconfounder}~\cite{bica2020time}. In their proposed work, the authors focus on addressing a specific type of hidden confounder which they refer to as multi-cause hidden confounders. The Time Series Deconfounder builds on a factor model to learn the distribution of treatments over time. By leveraging the dependence between multiple treatment options at each given time step, the factor model infers substitutes for unobserved confounders at each time step. The assumption of sequential strong ignorability is relaxed to sequential single strong ignorability where the assumption is limited to no hidden single-cause confounders~\cite{bica2020time}. The Times Series Deconfounder can be applied to the datasets before passing the deconfounded data to other outcome estimation models such as RMSM~\cite{lim2018forecasting} or CRN~\cite{bica2020estimating}. 

While the results show great promise, the Time Series Deconfounder only works when there are multiple treatment options and fails when there's a single treatment option at each time step. This limitation of the Time Series Deconfounder is related to its use on the dependence between multiple treatment options to infer substitutes of hidden confounders.  \textit{Sequential Deconfounder}, on the other hand,  is a method that deconfounds time-series data for ITE by fitting a  Gaussian Process (GP) latent variable model to capture any sequential dependence between the assigned treatments~\cite{hatt2021sequential}, without the limitation of depending on multiple treatments. The GP-based latent variable model aims to control for the substitutes and uses them in conjunction with outcome estimation models such as RMSM~\cite{lim2018forecasting} or CRN~\cite{bica2020estimating} to estimate treatment effects over time. While the Sequential Deconfounder does not require multiple treatments at each time step, it requires a special case of the ignorability assumption, which they refer to as \textit{Time-Invariant Unobserved confounding}~\cite{hatt2021sequential}. This assumption imposes a requirement that the hidden confounder is the same random variable at each time step.

\subsubsection{Noisy Proxies}
Most aforementioned studies utilize latent models to infer the hidden confounder in time-series ITE by capturing sequential dependencies. Recently, Deconfounding Temporal AutoEncoder (DTA) is an autorencoder-based model that utilizes noise proxies as an alternative to latent factor models to learn hidden embeddings that resemble the true hidden confounders~\cite{kuzmanovic2021deconfounding}. The main assumption in DTA builds on the fact that the observed covariates are not necessarily true confounders and assumes that the observed covariates are noisy proxies of the true confounders. DTA aims to learn a hidden embedding for which the ITE is the same when hidden confounders are present and when Sequential Strong Ignorbaility applies. To do so, DTA optimizes over a special loss that is referred to as a cause regularization penalty to yield outcomes and treatment assignments that are conditionally independent for each hidden embedding~\cite{kuzmanovic2021deconfounding}.

\section{Datasets and Evaluation}
Most studies found in the literature make use of simulated datasets to evaluate their methods for ITE estimation from time-series data. Unlike real-world data, where only the factual outcome is observed, simulations provide ground truths for all potential outcomes. Some simulated datasets used in the literature \cite{bica2020time, hatt2021sequential, kuzmanovic2021deconfounding} do not aim to mimic specific medical scenarios; instead, they are based on purely mathematical modeling of time-varying covariates, hidden confounders, treatments, and outcomes. In contrast, models such as the pharmacokinetic-pharmacodynamic (PK-PD) model by Geng et al. \cite{Geng2017}, which simulates cancer dynamics, strive to provide a realistic perspective on actual medical processes. Simulated "observational" cancer growth data, derived from the PK-PD model, is widely utilized in the literature \cite{lim2018forecasting, bica2020estimating, li2021g, melnychuk2022causal} for evaluation purposes. The model has been adapted to simulate the change in tumor volume over time under the influence of different treatment options, such as chemotherapy and radiation. Time-dependent confounding can be incorporated into the model by expressing the probabilities of administering chemotherapy and radiation as a function of tumor size \cite{lim2018forecasting}. 

In addition to the PK-PD model, Li et al. \cite{li2021g}  evaluate the performance of G-Net on longitudinal data, simulated using Heldt et al.'s \cite{Heldt2010-vz} CVSim. CVSim provides a mechanistic model of the human cardiovascular system and enables the simulation of trajectories of outcome parameters such as the mean arterial pressure (MAP) or the central venous pressure (CVP) under interventions such as different administration strategies of fluid or vasopressors. 

Moreover, Melnychuk et al. \cite{melnychuk2022causal} evaluate their Causal Transformer on both semi-synthetic and real-world datasets of ICU patient trajectories that are based on the MIMIC-III dataset \cite{Johnson2016}. For their semi-synthetic data, they combine real-world covariates from the MIMIC-extract by Wang et al. (2020) \cite{Wang2020} with simulated trajectories of control outcomes. Treated outcomes are obtained by simulating synthetic binary treatments, incorporating confounding, and applying those treatments to the control outcomes \cite{melnychuk2022causal}. For their experiments on real-world data, they use the same patient's covariates from MIMIC-III again and consider the effect of vasopressors and mechanical ventilation on blood pressure. However, since counterfactual outcomes are not available for real-world data, they can only report the performance in predicting the factual outcomes.

The fundamental problem of causal inference and the resulting reliance on (semi-) simulated data sets for comprehensive validation poses a major challenge to developing models for ITE estimation. Although models such as CVSim or the PK-PD model can offer valuable insights from a medical standpoint, their data generation process is less complex with simple assumptions, which could result in lower performance when compared to real-world applications. 

\section{Future Outlook}
\subsection{Irregular Sampling and Missingness}
All of the studies included in this work have shown great promise in estimating ITE in discrete time-series data. Transforming multi-variate time-series data to one with discrete time steps requires covariate alignment. Since most time-series covariates are measured at irregular time-steps where treatment is also administered at various time-steps,  missing values are inevitably produced in the data. Despite the increasing number of proposed approaches for estimating ITEs and deconfounding the data, none of the previous works have investigated the impact of missing values on counterfactual predictions and model performance. With a wide variety of methods proposed for handling time-series missingness with various underlying assumptions on the nature of missingness~\cite{fortuin2020gp,cao2018brits,du2023saits}, we believe that new research directions should investigate the impact of such assumptions on the ITE assumptions and models. This is particularly important when data missingness can be related to the underlying patient's health state, which can lead to a lack of follow-up, making it relevant patient information for the ITE estimation. 
\subsection{ITE and Dynamic Reinforcement Learning}
Another line of work that receives increased attention from the ITE research community is that of dynamic reinforcement learning (DRL)-based treatment regimes. While the objectives may appear similar, DRL and ITE have different underlying assumptions that are important to note. For instance, ITE makes strong assumptions about the nature of the data concerning confounders and treatment effects, which is not the case in DRL. DRL also assumes a Markovian state-data generation mechanism~\cite{komorowski2018artificial}, which is not necessarily required in ITE. Understanding the differences in the related work in literature can help one in understanding the limitations of integrating emerging works into real-world problems.

\printbibliography
\end{document}